\documentclass{article}

\usepackage{PRIMEarxiv}

\usepackage[utf8]{inputenc} % allow utf-8 input
\usepackage[T1]{fontenc}    % use 8-bit T1 fonts
\usepackage{hyperref}       % hyperlinks
\usepackage{url}            % simple URL typesetting
\usepackage{booktabs}       % professional-quality tables
\usepackage{amsfonts}       % blackboard math symbols
\usepackage{nicefrac}       % compact symbols for 1/2, etc.
\usepackage{microtype}      % microtypography
\usepackage{lipsum}
\usepackage{fancyhdr}       % header
\usepackage{graphicx}       % graphics
\graphicspath{{media/}}     % organize your images and other figures under media/ folder
\usepackage{algorithm} % For algorithm floating environment
\usepackage{algpseudocode} % For pseudocode notation
\usepackage{algorithm}
\usepackage{algpseudocode}
\usepackage{amsmath}

\title{Cooling-Guided Diffusion Model for Battery Cell Arrangement
\thanks{\textit{\underline{Citation}}: 
\textbf{Authors. Title. Pages.... DOI:000000/11111.}} 
}

\author{
  Nicholas Sung\thanks{Address all correspondence to this author.}, Faez Ahmed \\
  Department of Mechanical Engineering \\
  Massachusetts Institute of Technology \\
  Cambridge, MA\\
  \texttt{\{nicksung, faez\}@mit.edu} \\
   \And
  Zheng Liu, Pingfeng Wang \\
  Department of Industrial and Enterprise Systems Engineering \\
  University of Illinois Urbana-Champaign \\
  Urbana, Champaign, IL\\
  \texttt{\{zhengl6, pingfeng\}@illinois.edu} \\
}

\begin{document}

\maketitle    

%%%%%%%%%%%%%%%%%%%%%%%%%%%%%%%%%%%%%%%%%%%%%%%%%%%%%%%%%%%%%%%%%%%%%%
\begin{abstract}
{\it 
% Optimizing the design of cells within a battery pack significantly enhances the cooling performance and efficiency of battery thermal management systems. However, the conventional design process, which requires iterative optimization and performance evaluation, is slow and dependent on initial guesses. We propose an alternative approach that uses Generative AI with a cooling-guided diffusion model, aimed at generating feasible cell layouts with optimized cooling efficiency. We used a parametric denoising diffusion probabilistic model (DDPM) with both classifier guidance and cooling guidance in the sampling process. By utilizing position-based classifier-guided sampling, our approach generates feasible cell layouts with optimized cooling paths. This, combined with cooling-guided sampling, reduces the maximum temperature of the cells. We compared the quality of generated layouts by cooling-guided diffusion model with two state-of-the-art models, Tabular Denoising Diffusion Probabilistic Model (TabDDPM) and Conditional Tabular GAN (CTGAN). We found that the cooling-guided diffusion model achieves an overall quality that is 5 times superior to TabDDPM and 66 times superior to that of CTGAN when considering feasibility, diversity, and cooling efficiency metrics. This substantial advancement underscores the effectiveness of our proposed model in optimizing battery cell layouts for enhanced cooling efficiency, paving the way for more efficient and reliable battery thermal management systems.
Our study introduces a Generative AI method that employs a cooling-guided diffusion model to optimize the layout of battery cells, a crucial step for enhancing the cooling performance and efficiency of battery thermal management systems. Traditional design processes, which rely heavily on iterative optimization and extensive guesswork, are notoriously slow and inefficient, often leading to suboptimal solutions. In contrast, our innovative method uses a parametric denoising diffusion probabilistic model (DDPM) with classifier and cooling guidance to generate optimized cell layouts with enhanced cooling paths, significantly lowering the maximum temperature of the cells. By incorporating position-based classifier guidance, we ensure the feasibility of generated layouts. Meanwhile, cooling guidance directly optimizes cooling-efficiency, making our approach uniquely effective. When compared to two advanced models, the Tabular Denoising Diffusion Probabilistic Model (TabDDPM) and the Conditional Tabular GAN (CTGAN), our cooling-guided diffusion model notably outperforms both. It is five times more effective than TabDDPM and sixty-six times better than CTGAN across key metrics such as feasibility, diversity, and cooling efficiency.  This research marks a significant leap forward in the field, aiming to optimize battery cell layouts for superior cooling efficiency, thus setting the stage for the development of more effective and dependable battery thermal management systems.
}
\end{abstract}

\textbf{Keywords:} generative artificial intelligence; denoising diffusion probabilistic models; battery thermal management system, battery active cooling
%%%%%%%%%%%%%%%%%%%%%%%%%%%%%%%%%%%%%%%%%%%%%%%%%%%%%%%%%%%%%%%%%%%%%%
% \begin{nomenclature}
% \entry{A}{You may include nomenclature here.}
% \entry{$\alpha$}{There are two arguments for each entry of the nomemclature environment, the symbol and the definition.}
% \end{nomenclature}

% The spacing between abstract and the text heading is two line spaces.  The primary text heading is  boldface in all capitals, flushed left with the left margin.  The spacing between the  text and the heading is also two line spaces.

%%%%%%%%%%%%%%%%%%%%%%%%%%%%%%%%%%%%%%%%%%%%%%%%%%%%%%%%%%%%%%%%%%%%%%
\section*{INTRODUCTION}
% Zheng
% -why battery cooling is important
% -cooling system
% -what has been done in battery cooling design (brief review)
% -design space
% -contribution: simulation

Electric vehicles (EVs) are becoming popular aided by the advancement of lithium-ion battery technology allowing longer range and lesser charging time. At the same time, safety concerns and range anxiety have become significant obstacles for EVs~\cite{jung2015displayed}. One key solution is the battery thermal management system (BTMS)~\cite{kim2019review}. An effective BTMS ensures the safety of the battery pack while boosting the performance of the battery pack~\cite{bibin2020review}. Various BTMSs have been applied for battery packs, including air cooling, liquid cooling, refrigerant direct cooling, phase change material-based cooling, heat pipe-based cooling, and thermoelectric element-based cooling~\cite{lu2020research}.

Among different cooling methods, direct liquid cooling has demonstrated the ability to attain high heat transfer rates, thanks to the direct contact between cells and the coolant~\cite{sundin2020thermal}. The cell-to-pack (CTP) approach has also been widely applied in electric vehicles. CTP omits the cell module assembly, and as a result, it improves the volume utilization rate over 15\%~\cite{wang2021experimental}. However, CTP requires many batteries to be positioned appropriately within the battery pack to utilize the space efficiently~\cite{liu2023generative}. Especially when the battery pack is discharged at a high C-rate, typical for most high-performance electric vehicles. During this process, the battery cells generate a lot of heat, increasing the temperature. The effective battery cell layout can help with thermal management, eliminating the risk of thermal runaway and extending battery cycle life. Due to the complexity of the battery pack with immersion cooling, finding the optimal battery cell layout using only the experimental method is highly challenging. With the help of multiphysics finite element (FE) simulations, the data generation speed is greatly improved. However,  traditional trial and error approaches still cannot discover the optimal battery cell layout even with the help of FE simulations~\cite{li2023machine}. Especially for designing the layout of battery cells with immersion cooling, since the layout of battery cells has many combinations. Thus, we propose using Generative AI methods to learn from a collection of configurations and propose new configurations.
Thus, a cooling-guided diffusion model is adopted to further improve battery thermal management systems efficiency and battery pack performance by optimizing battery cell layout.

This work proposes the use of generative artificial intelligence to explore a broader range of battery layout configurations beyond traditional design methods. A sophisticated generative artificial intelligence model built for battery pack (BP) design should be able to generate BP design with feasible and diverse cell layouts with high cooling efficiency. This study highlights utilizing a cooling-guided denoising diffusion probabilistic model (DDPM) to efficiently generate configurations for battery cell arrangements that are feasible and optimized for cooling efficiency. 
This is achieved by training a classifier to distinguish between feasible and infeasible cell layouts and employing a surrogate model to predict the maximum cell temperature.
By incorporating both the classifier and surrogate model into the sampling process, the DDPM can generate a feasible BP design with optimized cooling efficiency. The principal contributions of this paper are summarized as follows:

\begin{enumerate}
  \item Developed a novel initialization method for cell configuration sampling, which employs random placements and a repulsive force model to generate over 100,000 unique cell layout configurations for battery packs.
  %A new initialization method for sampling non-overlapping cell configurations is presented for dataset generation. It begins with random placements within a constrained domain. This is complemented by a repulsive force model to adjust cell positions iteratively, encouraging a broad exploration of potential layouts.
  \item Implemented and validated a simplified simulation model that accelerates data generation by over 90\%, simplifying the battery discharging process into a dynamic heat source model based on high-fidelity electrothermal modeling.
  %The simulation model simplified the battery discharging process to a dynamic heat source model based on the high-fidelity electrothermal model, while the accuracy is still preserved.
  \item Trained a classifier on a dataset of 1.4 million layouts to accurately identify feasible battery cell layouts with an F1 score of 0.91, facilitating the efficient guidance of the generative model towards feasible designs.
  % A classifier, trained on a dataset of 1.4 million layouts, was introduced to learn the spatial relations between cells and guide the sampling process
  \item Utilized a synthetic minority over-sampling technique to address data skewness, significantly enhancing the performance of a surrogate model trained on a balanced dataset of 10,000 configurations, improving the prediction accuracy of cell maximum temperatures for better cooling efficiency.
  \item Demonstrated the superior performance of the cooling-guided diffusion model, which outperforms two state-of-the-art models, achieving a Composite Quality Index (CQI) that is 5 times higher than the Tabular Denoising Diffusion Probabilistic Model (TabDDPM) and 66 times higher than the Conditional Tabular GAN (CTGAN), showcasing its effectiveness in generating feasible and thermally efficient battery cell layouts.
  %SMOGN library was adapted to the skewed dataset, which markedly improved the surrogate model’s performance, especially in under-represented regions of the target variable distribution, demonstrating its utility beyond conventional applications.
\end{enumerate}

This paper is organized as follows: Section 2 explains the details of creating a comprehensive dataset for various BP designs. Section 3 illustrates the development of high-fidelity FE models for analyzing immersion cooling systems for the battery pack. Section 4 delves into the development and detailed evaluation of the cooling-guided denoising diffusion probabilistic model (DDPM) built for optimizing battery configurations. Section 5 compares the feasibility, diversity, and cooling efficiency of battery layouts generated by cooling-guided DDPM, TabDDPM~\cite{kotelnikov2023tabddpm} and Conditional Tabular GAN~\cite{ctgan}. Finally, Section 6 concludes key methods and findings. 

%%%%%%%%%%%%%%%%%%%%%%%%%%%%%%%%%%%%%%%%%%%%%%%%%%%%%%%%%%%%%%%%%%%%%%
\section*{CREATING THE BATTERY PACK DESIGN DATASET}
\begin{figure}[t]
\begin{center}
% Adjust the width of the image to suit your needs. For example, \textwidth.
\includegraphics[width = 0.4\textwidth]{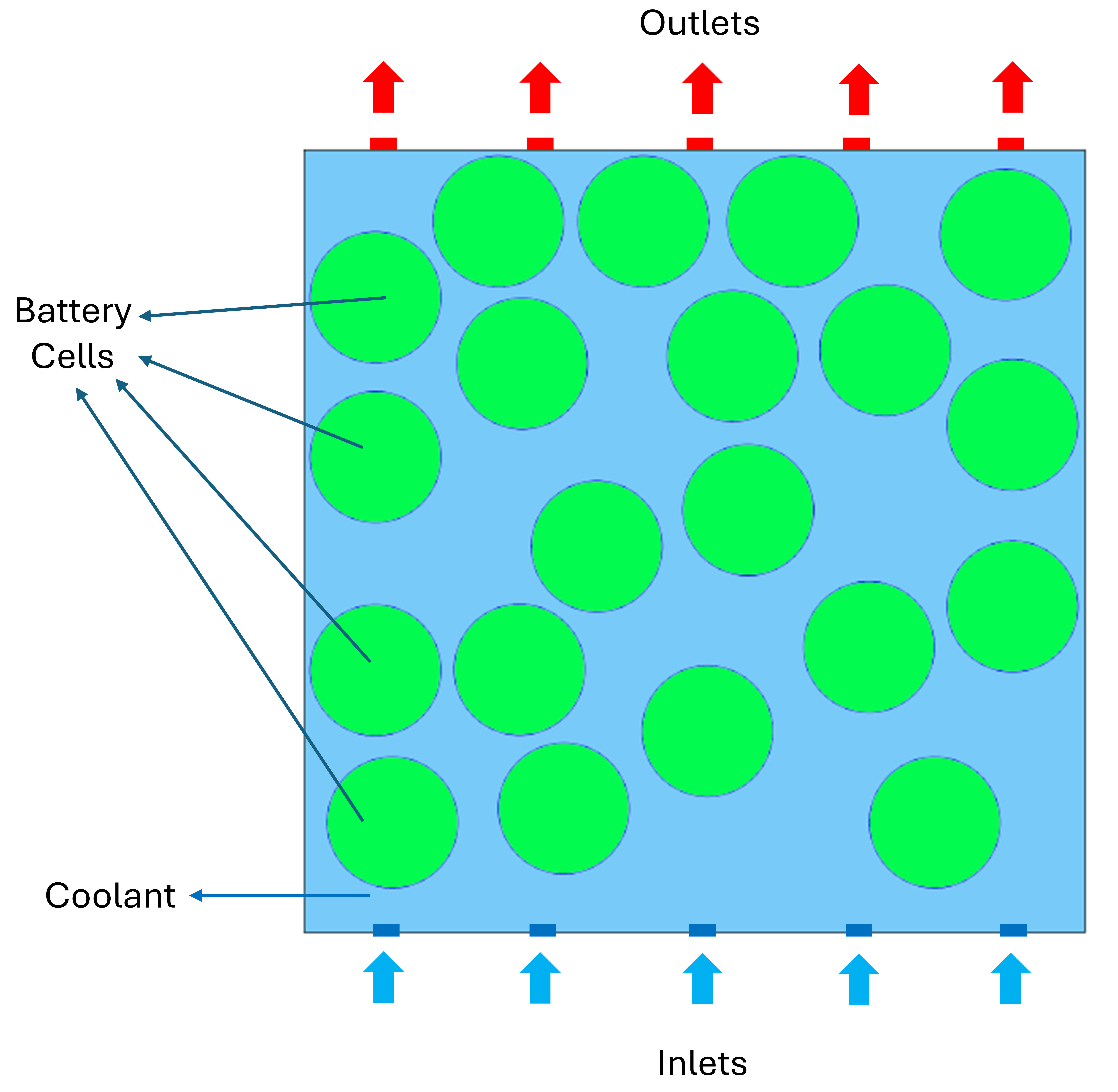}
\end{center}
\caption{SCHEMATIC OF IMMERSION COOLING FLOW IN A BATTERY PACK CONFIGURATION}
\label{figure_cells} 
\end{figure}

This section describes the method used to generate initial configurations of battery cells within a square battery pack. As shown in Figure ~\ref{figure_cells}, the battery pack contains 20 cells, a design informed by the literature \cite{wang2020cooling}. Specifically, the pack utilizes the 2170 battery cell, known for its application in Tesla Model 3 production, featuring a diameter of 21 mm. The pack itself is encased within a dimensionally precise frame measuring 125 by 125 mm. Notably, it is equipped with five inlets and five outlets, each extending 2 mm in width, to facilitate optimal cooling and performance.With the adoption of immersion cooling, the arrangement of battery cells within the pack offers a multitude of configurations. These varying layouts, despite maintaining a consistent coolant flow rate (and thus the same cooling energy expenditure), can lead to significant differences in the maximum temperatures reached by the cells. Therefore, identifying an optimal layout for the battery cells is crucial for improving the overall performance of the battery pack, primarily by minimizing the maximum temperature observed among the cells.

Given that the exploration of optimal battery cell layouts is an unexplored domain, our initial step involves creating an extensive dataset focused on battery pack designs. Generating the battery pack design dataset aims to position non-overlapping cylindrical battery cells within the constrained domain efficiently. The process begins with the algorithm randomly initializing cell positions within the constrained domain. This ensures a diverse set of starting configurations, which is crucial for exploring a wide range of potential layouts. Subsequently, the algorithm employs a repulsive force model to adjust the positions of the cells iteratively. When cells are detected to be closer than a minimum distance to each other, they are subjected to forces that simulate physical repulsion, pushing them apart to prevent overlap and ensure a minimum spacing. This methodology is encapsulated in in Algorithm 1.

\begin{algorithm}
\caption{Simulation of Repulsive Forces among Cells}
\begin{algorithmic}[1]
\State Initialize random points within domain bounds for all circles
\For{each iteration}
\For{each unique pair of circles $(i, j)$}
\State Calculate distance $d$ between circles $i$ and $j$
\If{$d > 0$}
\State Calculate overlap $O = 2 \times (radius + min_distance) - d$
\If{$O > 0$}
\State Force $F \propto O$
\State Apply $F$ to move circles $i$ and $j$ away from each other
\EndIf
\EndIf
\EndFor
\State Ensure all circles stay within domain bounds
\EndFor
\State \textbf{return} positions of circles
\end{algorithmic}
\end{algorithm}

A key aspect of this process involves exploring configurations under varying minimum distance constraints between cells. Specifically, we investigate minimum distances ranging from 1.0 mm to 1.9 mm in intervals of 0.1 mm. For each specified minimum distance, the algorithm generates 10,000 distinct configurations. By employing this iterative and exploratory approach, we generated 100,000 unique configurations.

Next, to prepare data for the simulation detailed in Section 3, we faced the challenge of the 100,000 configurations generated being too extensive, rendering the simulation process overly time-consuming. To address this, we employed a sampling approach, selecting a more manageable subset without compromising the diversity of the data. The sampling was achieved by first calculating the Intersection over Union (IoU) for each pair of configurations to quantify their spatial overlap, leveraging IoU's effectiveness in measuring similarity for spatial data. Next, given that k-medoids clustering requires a dissimilarity measure, we convert IoU values into distances by subtracting them from 1, thus inversely relating similarity to distance. This distance matrix forms the basis for k-medoids clustering, a method chosen for its ability to use real data points (medoids) as cluster centers, thereby ensuring that the selected representatives are actual configurations from the dataset. Combining IoU for similarity assessment and k-medoids for 500 clusters based on derived distances, we systematically reduced the configurations for each minimum distance category from 10,000 to 500 configurations. More importantly, this approach also ensured a broad representation of layout variations, enabling efficient simulation and effective training of the surrogate model with a significantly reduced computational load. Consequently, we have a total of 5,000 unique configurations to proceed through the simulation model, forming the surrogate model's training dataset.
%%%%%%%%%%%%%%%%%%%%%%%%%%%%%%%%%%%%%%%%%%%%%%%%%%%%%%%%%%%%%%%%%%%%%%
\section*{SIMULATION}
The simulation model is based on experimental testing of 2170 cylindrical cells with immersion cooling, as described by Liu et al.~\cite{liu2023control}. From the experiment, the battery cell's open-circuit voltage ($OCV$) of state of charge ($SoC$) at reference temperature ($T_{ref}$) was obtained. The data was imported to COMSOL Multiphysics software for the calculation of $OCV$ at a certain temperature $E_{OCV}(SoC,T)$.

\begin{equation}
E_{OCV}(SoC,T)=E_{OCV}(SoC,T_{ref})+(T-T_{ref})dE\label{eq}
\end{equation}
where $ dE $ is the voltage derivative, and it can be calculated by $OCV$ from the test, $E_{OCV}(SoC,T_{ref})$, and temperature derivative of open circuit voltage.
\begin{equation}
dE=\frac{dE_{OCV}(SoC,T_{ref})}{dT}\label{eq}
\end{equation}

The cell voltage ($E_{cell}$) can be obtained using the $OCV$ curve, ohmic overpotential ($\eta_{act}$), and activation overpotential ($\eta_{act}$).
\begin{equation}
E_{cell}=E_{OCV}(SoC,T)+\eta_{IR}+\eta_{act}\label{eq}
\end{equation}

In the simulation, the SOC of the battery was calculated through time evolution based on the battery cell capacity ($Q_{cell,0}$) and current ($I_{cell}$):
\begin{equation}
\frac{\partial SoC}{\partial t} = \frac{I_{cell}}{Q_{cell,0}}\label{eq}
\end{equation}

The ohmic overpotential (\(\eta_{IR}\)) was obtained from the ohmic
overpotential at 1C (\(\eta_{IR,1C}\)), with the current at 1C
(\(I_{1C}\)):

\begin{equation}
\eta_{IR} = \eta_{IR,1C}\frac{I_{cell}}{I_{1C}}\label{eq}
\end{equation}

The activation overpotential (\(\eta_{act}\)) on electrode surfaces can
be calculated as:

\begin{equation}
\eta_{act} = \frac{2RT}{F}{asinh}\left( \frac{I_{cell}}{2J_{0}I_{1C}}\right)\label{eq}
\end{equation}
where \(R\) denotes the molar gas constant, \(F\) is Faraday's constant,
a \(J_{0}\) is the dimensionless charge exchange current. Only the
active material layer is defined in the lumped battery module for the
battery cell. Heat transfer in the solids and fluids model is adopted,
and the temperatures of other components are calculated by solving for
heat transfer in solid, with consideration of the specific heat capacity
(\(C_{p}\)), density (\(\rho\)) of each material, absolute temperature
(\(T)\), the velocity vector of translational motion (\(\mathbf{u}\)),
heat flux by conduction and radiation
\textbf{(}\(\mathbf{q}, \mathbf{q}_{r}\)), coefficient of
thermal expansion (\(\alpha\)), second Piola-Kirchhoff stress tensor
(\(\mathbf{S}\)), and additional heat sources (\(\mathbf{Q}\)).

\begin{equation}
\rho C_{p}\left( \frac{\partial T}{\partial t} + \mathbf{u} \cdot \nabla T \right) + \nabla \cdot \left( \mathbf{q} + \mathbf{q}_{\mathbf{r}} \right) = - \alpha T:\frac{d\mathbf{S}}{dt} + \mathbf{Q}\label{eq}
\end{equation}

We consider the dielectric immersion cooling system. 3M Novec 7300
dielectric fluid has a thermal conductivity of
0.063 $W/(m \cdot K)$~\cite{chen2020immersion} and a specific heat capacity of 880
$J/(kg \cdot K)$~\cite{yu2022novel}. The fluid domain heat transfer is
calculated using the energy equation by:

\begin{equation}
\begin{split}
\rho C_{p}\left( \frac{\partial T}{\partial t} + \mathbf{u} \cdot \nabla T \right) + \nabla \cdot \left( \mathbf{q} + \mathbf{q}_{\mathbf{r}} \right) 
& \\
= - \alpha_{p}T\left( \frac{\partial p}{\partial t} + \mathbf{u} \cdot \nabla p \right) + \mathbf{\tau}:\nabla\mathbf{u} + \mathbf{Q\ }\label{eq} 
\end{split}
\end{equation}
where \(\alpha_{p}\) is the coefficient of thermal expansion. The
boundary of the battery pack is thermally insulated. The inlet coolant
temperature and flow rates are set to be the same as those observed in
the experimental setup. Though simulation greatly improves the data-generating speed compared to experimental methods, the computational cost still makes it challenging to generate enough training data for the surrogate model. Thus, the model was simplified to accelerate the simulation.

\begin{figure}[t]
\begin{center}
% Adjust the width of the image to suit your needs. For example, \textwidth.
\includegraphics[width = 0.4\textwidth]{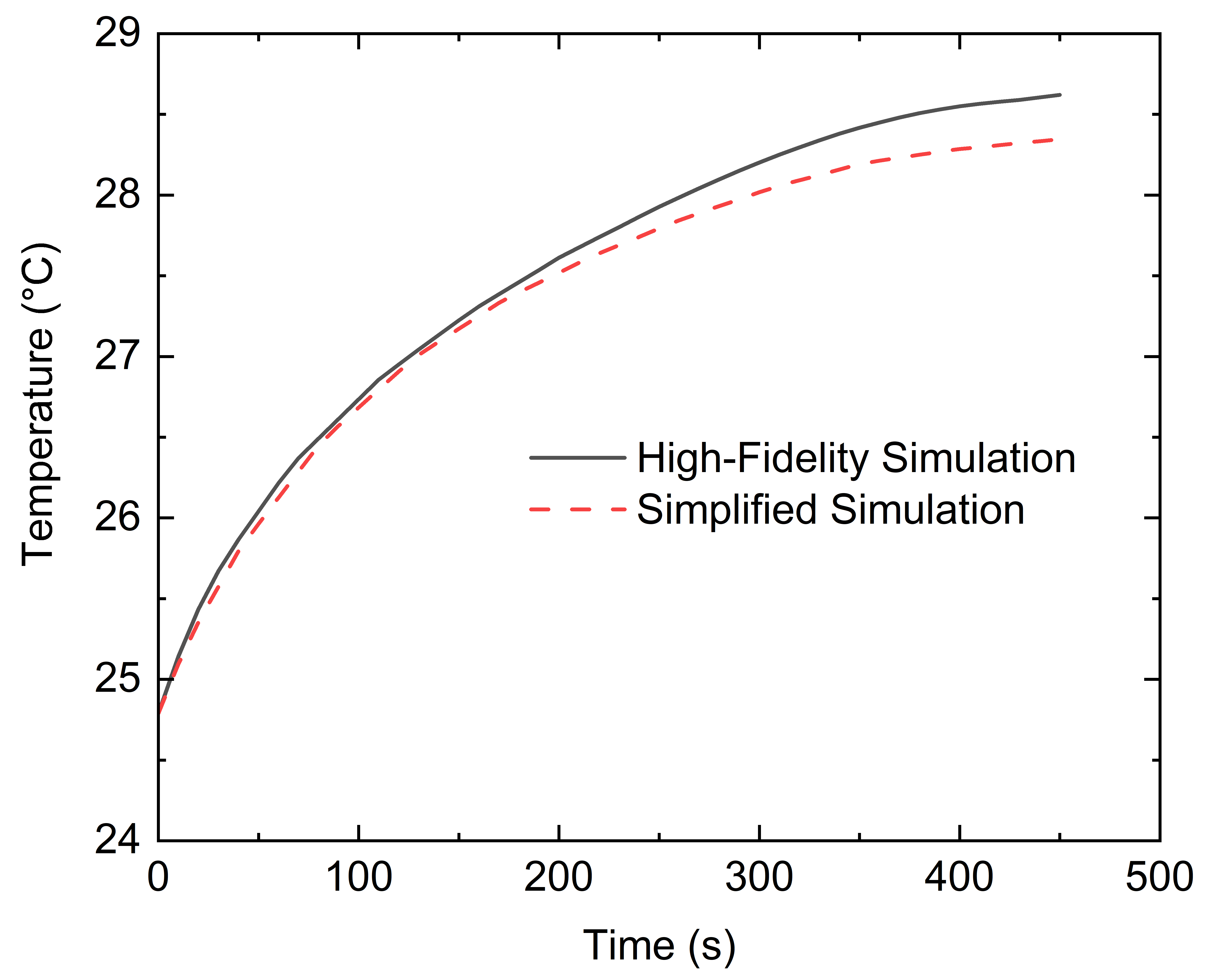}
\end{center}
\caption{TEMPERATURE TRAJECTORY: HIGH-FIDELITY VS. SIMPLIFIED SIMULATIONS OVER TIME}
\label{figure_sim} 
\end{figure}

Since only the 2C discharging process was investigated, the heat generated by battery cells can be considered a function of time. This function can be obtained from the high-fidelity model. Then, the simulation can be converted to a problem without calculating the battery's electrical properties. Furthermore, the battery cell is considered an entirety to simplify the simulation, though it has several layers. The 2C discharging process is from SoC = 100\% to SoC = 75\%. Also, we validated the simplified one-cell simulation result with the original high-fidelity model. Figure ~\ref{figure_sim} shows the accuracy of the battery's temperature is preserved with the simplified model while the computational cost is decreased by more than 90\%.

% \section*{PAPER NUMBER}

% ASME assigns each accepted paper with a unique number. Replace {\bf DETC98/DAC-1234} in the input file preamble (the location will be obvious) with the paper number supplied to you  by ASME for your paper.
%%%%%%%%%%%%%%%% begin figure %%%%%%%%%%%%%%%%%%%
\begin{figure*}[t]
\begin{center}
% Adjust the width of the image to suit your needs. For example, \textwidth.
\includegraphics[width = \textwidth]{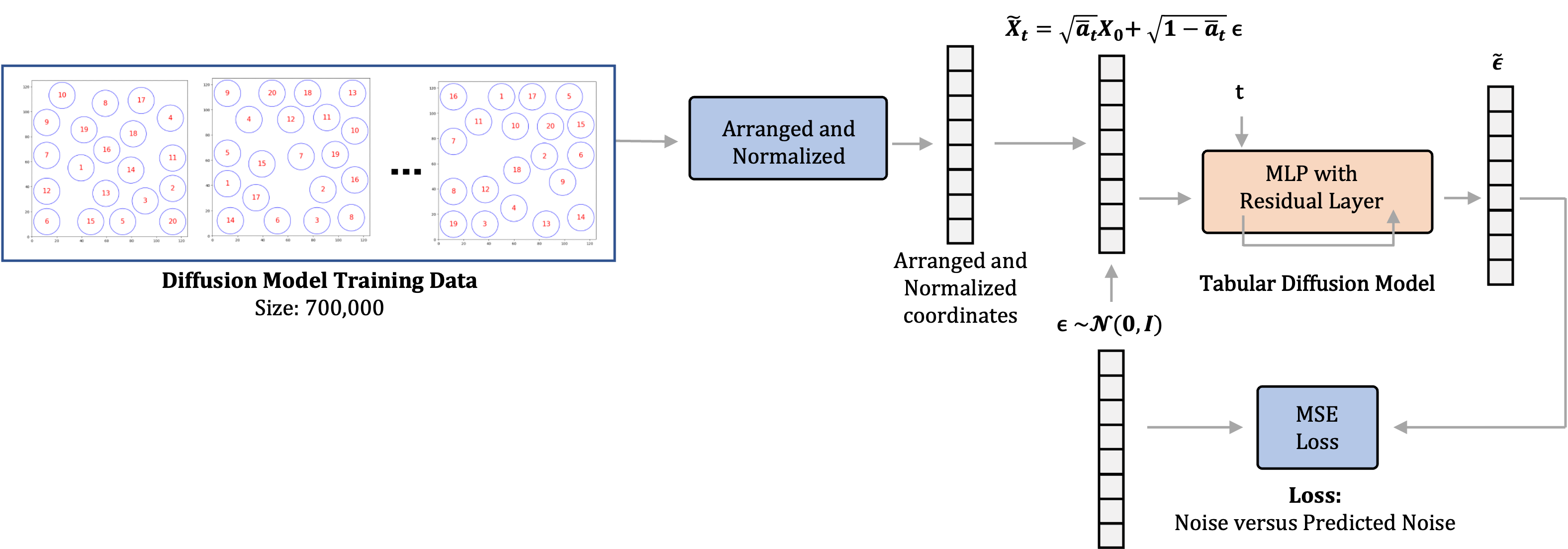}
\end{center}
\caption{OVERVIEW OF DDPM TRAINING}
\label{figure_training_DDPM} 
\end{figure*}
%%%%%%%%%%%%%%%% end figure %%%%%%%%%%%%%%%%%%%

%%%%%%%%%%%%%%%%%%%%%%%%%%%%%%%%%%%%%%%%%%%%%%%%%%%%%%%%%%%%%%%%%%%%%%
\section*{COOLING-GUIDED DDPM}
In this section, we start with an overview of how denoising diffusion probabilistic models are trained and sampled. Next, we explain how classifier guidance helps guide the sampling towards feasible layouts. Finally, we discuss the role of cooling guidance in directing the sampling towards layouts that are efficient in cooling.

\subsubsection*{DENOISING DIFFUSION PROBABILISTIC MODELS}
A denoising diffusion probabilistic model (DDPM) is a type of generative AI model designed to produce new data by progressively denoising a random input across multiple iterations, ensuring the generated data aligns with the statistical distribution of samples in the training data. In this paper, we will be using a specialized tabular DDPM , inspired by Kotelnikov et al., known as TabDDPM~\cite{kotelnikov2023tabddpm}. 

For the training dataset, the initial 100,000 unique configurations were subjected to processing that included mirroring and rotating, resulting in a total of 700,000 configurations. During training,  the cells are rearranged by sorting and grouping, starting from the one closest to the origin (bottom left), moving rightwards row by row. A linear transformation rescaled the bounds so that the domain is between 0 and 1. The DDPM is trained to predict the noise vector based on the timestep embedding and the partially noised vector. The loss for this prediction was calculated as the mean squared error (MSE) between the predicted noised vector and an actual noise vector. This MSE loss was then used to back-propagate through the DDPM, updating its weights and biases. This training process was executed for all 700,000 feasible coordinates over one thousand denoising time steps, utilizing random batches to effectively train the DDPM. This training process is shown in Figure ~\ref{figure_training_DDPM}. After training, the standard DDPM generates new coordinates by the sampling process which denoises a random Gaussian noise vector. The sampling algorithm is defined by Ho et al.~\cite{ho2020denoising} as shown in Algorithm 2.

% Here you would insert your Table 2. Replace this line with your actual table code.
\begin{algorithm}
\caption{Sampling algorithm for cooling-guided DDPM}
\begin{algorithmic}[1]
\State $X_T \sim N(0, I)$
\For{$t = T, \ldots, 1$}
\State $z \sim N(0, I)$ if $t > 1$, else $z = 0$
\State $X_{t-1} = \frac{1}{\sqrt{\alpha_t}} \left( X_t - \frac{1-\alpha_t}{\sqrt{1-\bar{\alpha}_t}} \varepsilon_{\theta}(X_t, t)) \right) + \sigma_t z$
\EndFor
\State \textbf{return} $x_0$
\end{algorithmic}
\end{algorithm}

%%%%%%%%%%%%%%%% begin figure %%%%%%%%%%%%%%%%%%%
\begin{figure*}[t]
\begin{center}
\includegraphics[width = 0.9\textwidth]{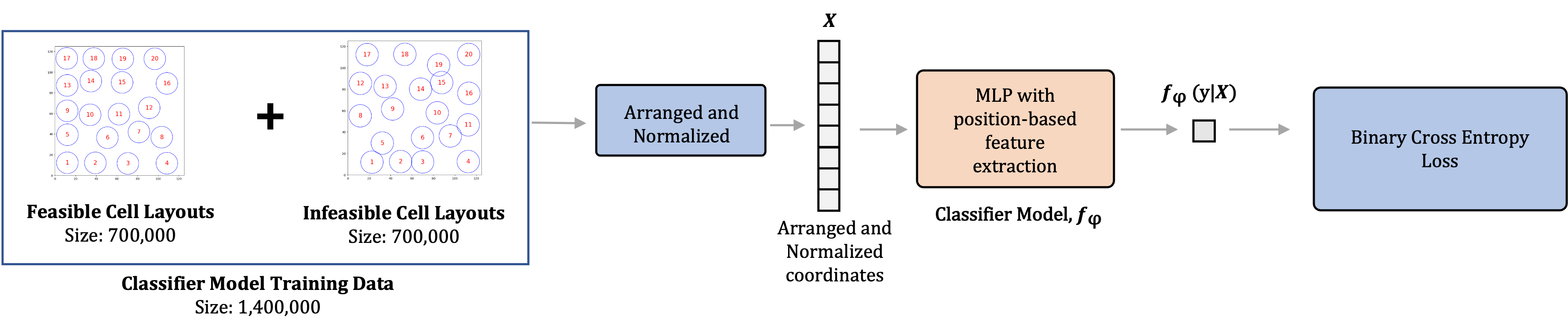}
\end{center}
\caption{OVERVIEW OF CLASSIFIER TRAINING}
\label{figure_training_classifier} 
\end{figure*}
%%%%%%%%%%%%%%%% end figure %%%%%%%%%%%%%%%%%%%

%%%%%%%%%%%%%%%% begin figure %%%%%%%%%%%%%%%%%%%
\begin{figure*}[t]
\begin{center}
% Adjust the width of the image to suit your needs. For example, \textwidth.
\includegraphics[width = \textwidth]{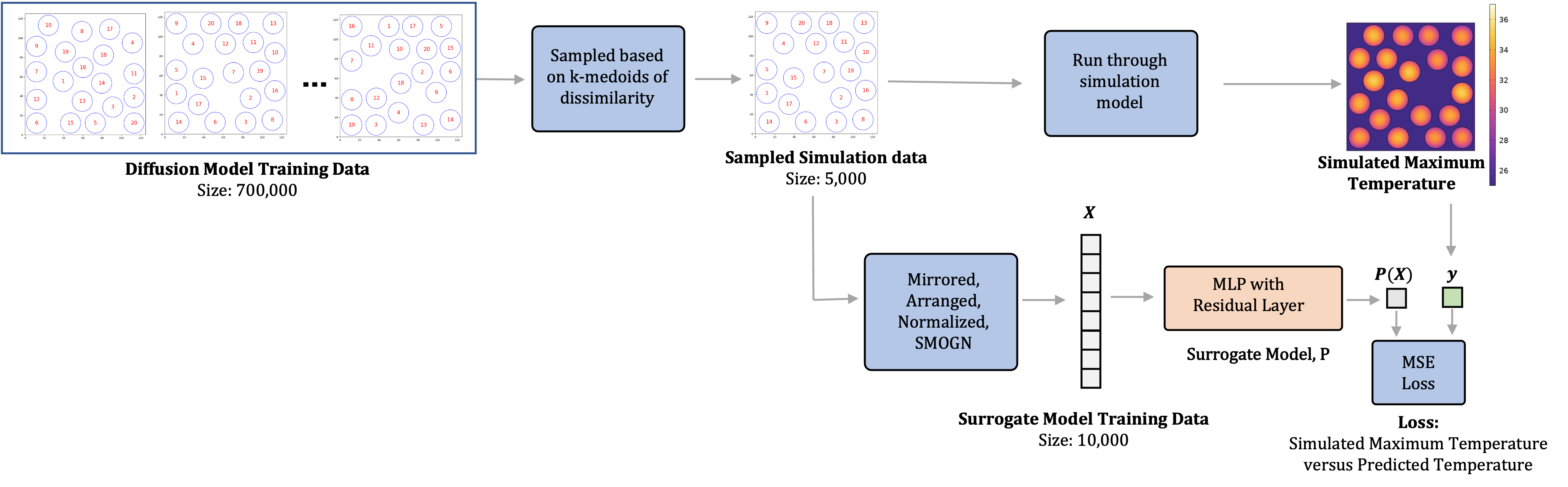}
\end{center}
\caption{OVERVIEW OF SURROGATE MODEL TRAINING}
\label{figure_training_regressor} 
\end{figure*}
%%%%%%%%%%%%%%%% end figure %%%%%%%%%%%%%%%%%%%

\subsubsection*{CLASSIFIER GUIDANCE FOR DIFFUSION MODEL}
We used a classifier to guide the generation towards feasible layouts. This is done by leveraging the gradients from a classifier that has been trained to distinguish between feasible and infeasible designs during the sampling process of a standard DDPM. Here, the classifier was trained to distinguish whether a layout is feasible, based on having non-overlapping cells that fit within the domain. We used 700,000 feasible layouts and 700,000 infeasible layouts (generated by an unconditioned DDPM) to assess design feasibility. The classifier as shown in Figure ~\ref{figure_training_classifier} is trained by leveraging geometric relationships between cells by computing the edge-to-edge distances between pairs of cells, aggregating these distances through a linear layer to perform binary classification based on the spatial arrangement of cells.  This classifier was successfully trained, achieving an F1 score of $\textit{0.91}$, indicating high precision and recall in identifying feasible layouts. 

After training the classifier, it is employed to guide the sampling towards feasible layouts. During each timestep of the sampling process, the gradient of the classifier's prediction for a given class, represented as  \( f_{\phi}(y|X_t) \), with respect to the parameterized design vector, \( X_t \), was calculated~\cite{dhariwal2021diffusion}. This computed gradient was multiplied by a hyperparameter, \( \gamma \), and was added to the sample during Step 4 of the DDPM sampling algorithm defined in Algorithm 2. A classifier-guided DDPM was created by replacing Step 4 with Equation \ref{eq_sampling_classifier}. 

\begin{align}
X_{t-1} &= \frac{1}{\sqrt{\alpha_t}} \left( X_t - \frac{\sqrt{1-\bar{\alpha}_t}\epsilon_{\theta}(X_t, t)}{\sqrt{1-\alpha_t}} \right) \nonumber \\
&\quad + \sigma_t(Z(1-\gamma)) + \gamma \nabla_{X_t}f_{\phi}(y|X_t) 
\label{eq_sampling_classifier}
\end{align}

%%%%%%%%%%%%%%%% begin figure %%%%%%%%%%%%%%%%%%%
\begin{figure*}[t]
\begin{center}
% Adjust the width of the image to suit your needs. For example, \textwidth.
\includegraphics[width = \textwidth]{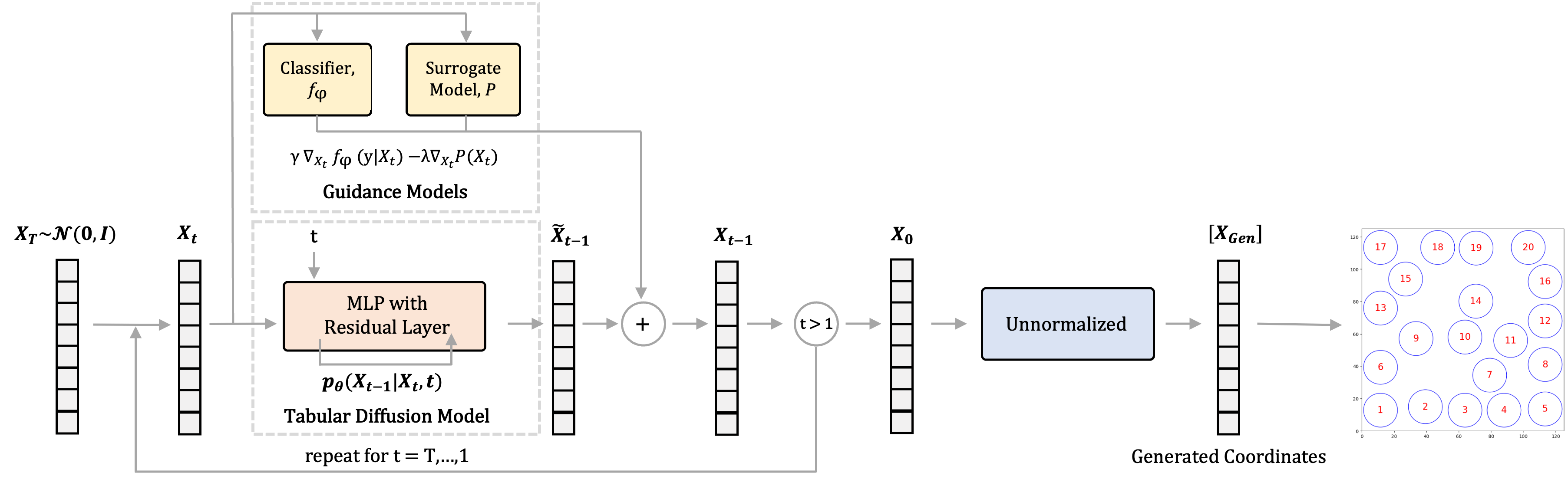}
\end{center}
\caption{OVERVIEW OF GUIDED DDPM SAMPLING}
\label{figure_sampling} 
\end{figure*}
%%%%%%%%%%%%%%%% end figure %%%%%%%%%%%%%%%%%%%

\subsubsection*{COOLING GUIDANCE FOR DIFFUSION MODEL}
\begin{figure}[t]
\begin{center}
% Adjust the width of the image to suit your needs. For example, \textwidth.
\includegraphics[width = 0.4\textwidth]{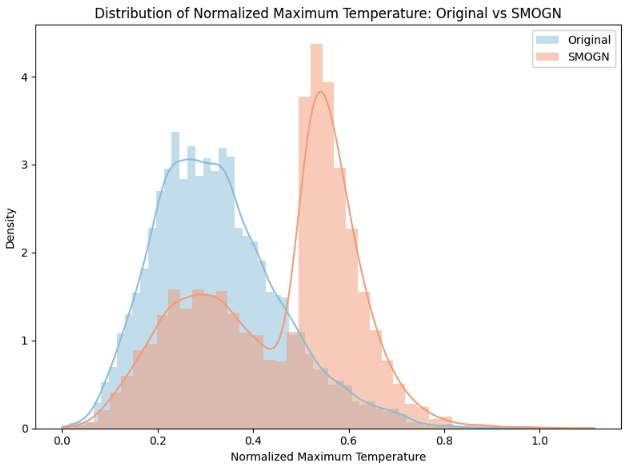}
\end{center}
\caption{ENHANCED SAMPLING IN SPARSE REGIONS BY SMOGN FOR NORMALIZED MAX TEMPERATURE DISTRIBUTION}
\label{figure_smogn} 
\end{figure}

Similar to the concept of classifier guidance, we built a surrogate model to estimate the cooling efficiency of a battery cell layout which will be employed to guide the generation of samples. From the initial set of 5,000 unique configurations we selected for simulation, we doubled the dataset to 10,000 viable battery cell layouts by mirroring each configuration's coordinates along the vertical axis. This process allowed us to enrich the dataset with variations, ensuring that each layout included its corresponding mirrored version, all while preserving the crucial parameter of maximum temperature. 
Due to the limited size and the skewed nature of the dataset, we employed the SMOGN (Synthetic Minority Over-sampling Technique for Regression with Gaussian Noise) method~\cite{branco2017smogn}. This method is specifically designed to address challenges in regression tasks where the distribution of target values is imbalanced. SMOGN applies a synthetic over-sampling technique that generates new samples closer to rare or extreme values in the continuous target space, thereby enriching the dataset and improving the model's ability to learn from under-represented regions of the target variable distribution. We could observe the improved and more balanced distribution in Figure ~\ref{figure_smogn}.

%%%%%%%%%%%%%%%% begin figure %%%%%%%%%%%%%%%%%%%
%%%%%%%%%%%%%%%% end figure %%%%%%%%%%%%%%%%%%%

The surrogate model was built with a residual network with two hidden layers with 256 nodes where the first hidden layer was added as a residual to the second hidden layer. Figure ~\ref{figure_training_regressor} illustrates the entire training process of the surrogate model. Given the constraints of a small dataset, we employed a k-folds cross-validation approach to maximize the utility of the data. This method allowed us to systematically split the dataset into five smaller sets to ensure that each dataset had a chance to be tested.  
We evaluated our dataset of 10,000 training samples before and after applying the SMOGN transformation. Initially, the model exhibited an average training $R^2$ score of 0.98, but a lower average test $R^2$ of 0.34, indicating a discrepancy between training and testing performance. After the SMOGN transformation, the training $R^2$ score remained stable at 0.98, while the test $R^2$ improved to 0.81. This improvement in test performance indicates that the SMOGN transformation significantly helped in enhancing the model's predictive accuracy on the normalized maximum temperature, balancing the model's ability to generalize, particularly for unseen data.

After training the surrogate model, it is used to guide the generation towards cell layouts with low maximum temperature. During each time step of the sampling process, the gradient of the surrogate models's prediction, represented as  \( P(X_t) \), with respect to the parameterized design vector, \( X_t \), was calculated~\cite{bagazinski2023shipgen}. This computed gradient was multiplied by a hyperparameter, \( \lambda \), and was added to the sample during Step 4 of the DDPM sampling algorithm defined in Algorithm 2. By integrating the gradients from both the surrogate model and the classifier, we derive the update equation for each time step. This is represented as Equation \ref{eq_sampling_overall} which is intended to replace Step 4 in Algorithm 2. The overall sampling process of the guided DDPM is shown in Figure ~\ref{figure_sampling}.

\begin{align}
X_{t-1} &= \frac{1}{\sqrt{\alpha_t}} \left( X_t - \frac{1 - \alpha_t}{\sqrt{1-\bar{\alpha}_t}}\epsilon_{\theta}(X_t, t) \right) \nonumber \\
&\quad + \sigma_t(Z(1 - \gamma)) + \gamma \nabla_{X_t}f_{\phi}(y|X_t) - \lambda \nabla_{X_t} P(X_t) \label{eq_sampling_overall}
\end{align}

%%%%%%%%%%%%%%%%%%%%%%%%%%%%%%%%%%%%%%%%%%%%%%%%%%%%%%%%%%%%%%%%%%%%%%
\section*{RESULTS}
In this section, we begin by outlining the metrics for evaluating the quality of generated cell layouts, with a focus on feasibility, diversity, and cooling efficiency. Then, we analyse the effect of hyperparameters, specifically \( \lambda \) and \( \gamma \), on the quality of generated cell layouts. Finally, we assess the effectiveness of the cooling-guided DDPM by comparing it to two state-of-the-art methods, Tabular Denoising Diffusion Probabilistic Model (TabDDPM)~\cite{kotelnikov2023tabddpm} and Conditional Tabular GAN (CTGAN)~\cite{ctgan}, in terms of the quality of the generated cell layouts.

\subsection*{METRICS FOR EVALUATING QUALITY OF GENERATED CELL LAYOUTS}
The quality of the generated layouts are assessed based on the following three metrics:
\begin{enumerate}
    \item \textbf{Feasibility}: Evaluated through the feasibility rate, \( f_r \), which measures the proportion of generated layouts that meet predefined constraints. It can be calculated as:
    \begin{equation}
    f_r = \frac{\text{Number of Feasible Layouts}}{\text{Total Number of Generated Layouts}}
    \end{equation}
    
    \item \textbf{Diversity}: Calculated by the average diversity score, \( D_s \), which uses the Intersection over Union (IoU) approach for each unique pair of feasible layout. The average diversity score is given by:
    \begin{equation}
    D_s = 1 - \frac{1}{\binom{n}{2}}\sum_{i=1}^{n-1}\sum_{j=i+1}^{n} \text{IoU}_{ij}
    \end{equation}
    where \( \text{IoU}_{ij} \) is the Intersection over Union for the \( i^{th} \) and \( j^{th} \) layout, and \( n \) is the total number of feasible layouts.
    
    \item \textbf{Cooling Efficiency}: Determined by the average normalized maximum temperature, \( \overline{T_{norm}} \), across all feasible layouts predicted by the surrogate model. This provides an evaluation of the cooling efficiency of the layouts and is calculated as:
    \begin{equation}
    \overline{T_{norm}} = \frac{1}{n}\sum_{k=1}^{n} \frac{T_{max} - P(X_k)}{T_{max} - T_{min}}
    \end{equation}
    where \( T_{max} \) and \( T_{min} \) are the maximum and minimum temperatures observed in the training data, \( P(X_k) \) is the predicted temperature for the \( k^{th} \) generated layout, and \( n \) is the total number of feasible layouts.
\end{enumerate}
Higher values of \( f_r \), \( D_s \), and \( \overline{T_{norm}} \) indicate better generated layouts. Specifically, a higher \( f_r \) reflects greater compliance with constraints, a higher \( D_s \) suggests a more diverse set of solutions, and a higher \( T_{norm} \) indicates a lower predicted temperature relative to the range, hence, better cooling efficiency.

To holistically evaluate the quality of the generated layouts, we introduce Composite Quality Index (CQI), which is the product of the three metrics:
\begin{equation}
\text{CQI} = f_r \times D_s \times \overline{T_{norm}}
\end{equation}

A higher value of CQI indicates that the generated layouts excel across all evaluated criteria: feasibility, diversity, and cooling efficiency. This composite metric serves as a comprehensive measure of layout quality.

%%%%%%%%%%%%%%%% begin figure %%%%%%%%%%%%%%%%%%%
\begin{figure*}[t]
\begin{center}
% Adjust the width of the image to suit your needs. For example, \textwidth.
\includegraphics[width = \textwidth]{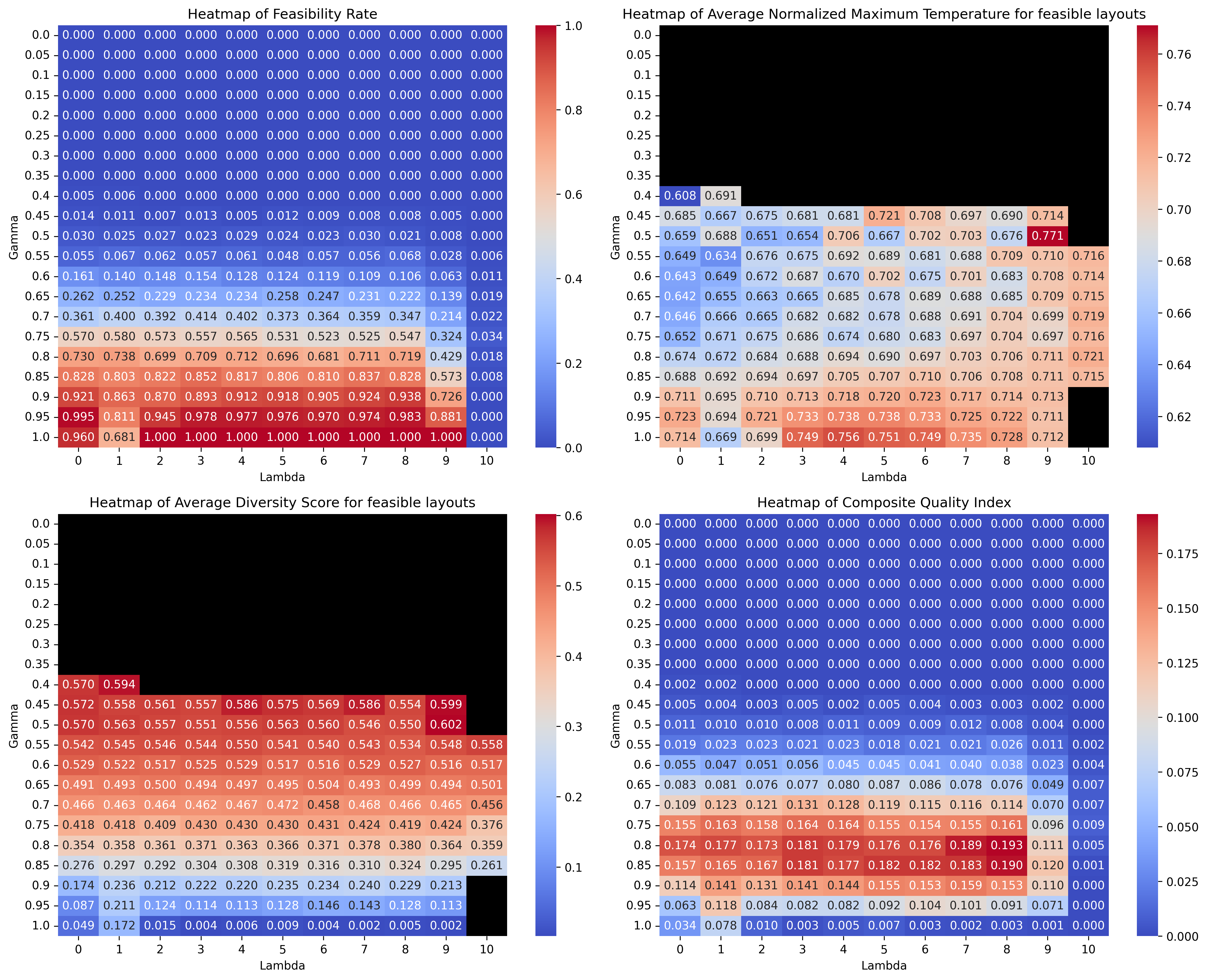}
\end{center}
\caption{HEAT MAP OF FEASIBILITY RATE, AVERAGE NORMALIZED MAXIMUM TEMPERATURE, AVERAGE DIVERSITY SCORE, AND COMPOSITE QUALITY INDEX WITH RESPECT TO \(\gamma\) AND \(\lambda\)}
\label{figure_heatmaps} 
\end{figure*}
%%%%%%%%%%%%%%%% end figure %%%%%%%%%%%%%%%%%%%

\subsection*{HYPERPARAMETER (\( \lambda \) and \( \gamma \)) EFFECT ON QUALITY OF GENERATED CELL LAYOUTS}

These hyperparameters are key in controlling the balance between feasibility, diversity, and cooling efficiency in the generated cell layouts. We adjust the hyperparameter \( \gamma \), associated with classifier's gradient to be from 0 to 1 with increments of 0.05. Similarly, \( \lambda \), associated with the surrogate model's gradient, varies from 0 to 10 in increments of 1. For each configuration of \( \lambda \) and \( \gamma \), we generated 1000 cell layouts and assessed their quality using metrics \( f_r \), \( D_s \), \( \overline{T_{norm}} \), and CQI. The outcomes of this evaluation, illustrating the effects of \( \lambda \) and \( \gamma \) on these four metrics, are visually represented through heat maps in Figure~\ref{figure_heatmaps}.

Based on the heat maps in Figure~\ref{figure_heatmaps}, we observe the following about the impact of the hyperparameters, \(\gamma\) (associated with feasibility classifier) and \(\lambda\) (associated with surrogate model for mean normalized temperature) on the three metrics:

\begin{enumerate}
   
  \item \textbf{Feasibility Rate}:
  \begin{itemize}
    \item There is a positive correlation between \(\gamma\) and the feasibility rate, with higher \(\gamma\) values resulting in a greater number of feasible solutions, reinforcing the role of \(\gamma\) in achieving feasibility.
    \item The feasibility rate also increases with \(\lambda\), especially at higher \(\gamma\) values, implying that the weight of the surrogate model contributes more significantly to the feasibility when the feasibility classifier is already given substantial consideration.
  \end{itemize}
  
    \item \textbf{Average Diversity Score}:
  \begin{itemize}
    \item The average diversity score appears to be relatively stable across various \(\gamma\) levels, suggesting that the diversity of feasible solutions is not strongly dependent on the weight given to the feasibility classifier.
    \item Conversely, an increase in \(\lambda\) tends to slightly decrease the average diversity score. This indicates a trade-off between optimizing for cooling efficiency and maintaining diversity among the feasible cell layouts..
  \end{itemize}
  
  \item \textbf{Average Normalized Maximum Temperature}:
  \begin{itemize}
    \item An increase in \(\gamma\) generally leads to a higher average normalized maximum temperature, indicating that higher weights on the feasibility classifier promote not only the feasibility but also the cooling efficiency of the generated layouts.
    \item The influence of \(\lambda\) on the average normalized maximum temperature is not distinctly linear or consistent across the range of \(\gamma\), suggesting a complex interaction between the feasibility and the regression objective that does not adhere to a simple trend.
  \end{itemize}
\end{enumerate}

These observations underscore the effect of the hyperparameters on the three metrics, suggesting that fine tuning \(\gamma\) and \(\lambda\) is required to balance feasibility, cooling efficiency, and diversity of solutions. In the context of optimizing the Composite Quality Index, which serves as a holistic measure of layout quality, a strategic selection of \(\gamma = 0.8\) and \(\lambda = 8\) is made, attributed to its superior CQI value. This configuration will be benchmarked against layouts generated by a CTGAN model to assess its effectiveness in generating high-quality layouts.

\subsection*{COMPARATIVE ANALYSIS OF CELL LAYOUTS GENERATED BY COOLING-GUIDED DDPM AND CTGAN}
For consistency, we employed the dataset of 700,000 feasible layouts (used for cooling-guided DDPM training) to train both TabDDPM and CTGAN. 
TabDDPM is a state-of-the-art unconditioned diffusion model for tabular data, which enhances data normalization through a quantile transformer, ensuring the training data follows a normal distribution for optimal generative performance~\cite{kotelnikov2023tabddpm}. This is unlike the cooling-guided DDPM which employs min-max normalization on the training data. We used the same denoising model for both cooling-guided DDPM and TabDDPM for a fair comparison.
CTGAN is a state-of-the-art framework, a composite of deep learning-driven synthetic data generators for single-table data, designed to capture and replicate the complexities of real data to produce synthetic datasets of high fidelity~\cite{ctgan}. A comparative assessment of the layouts generated by TabDDPM, CTGAN, and Cooling-Guided DDPM, is presented in Table \ref{table_quality_comparison}. 
%%%%%%%%%%%%%%%% begin table %%%%%%%%%%%%%%%%%%%%%%%%%%
% \usepackage{graphicx} % Load the package in the preamble if not already done

% \begin{table}[t]
% \caption{QUALITY COMPARISON OF LAYOUTS FROM COOLING-GUIDED DDPM, TABDDPM, and CTGAN}
% \label{table_quality_comparison}
% \resizebox{\columnwidth}{!}{%
% \begin{tabular}{lcccc}
% \hline
% Metric & Cooling-Guided DDPM & TabDDPM & CTGAN \\
% \hline
% Feasibility Rate (\( f_r \)) & \textbf{0.719} & 0.068 & 0.007 \\
% Diversity Score (\( D_s \)) & 0.380 & \textbf{0.801} & 0.617 \\
% Normalized Temperature (\( \overline{T_{norm}} \)) & \textbf{0.706} & 0.665 & 0.666 \\
% Composite Quality Index (CQI) & \textbf{0.1929} & 0.0362 & 0.0029 \\
% \hline
% \end{tabular}%
% }
% \end{table}

\begin{table}[t]
\caption{QUALITY COMPARISON OF LAYOUTS FROM COOLING-GUIDED DDPM, TABDDPM, AND CTGAN}
\label{table_quality_comparison}
\centering % Center the table
\small % Make the font of the table smaller
\begin{tabular}{lcccc}
\hline
Metric & CG DDPM & TabDDPM & CTGAN \\
\hline
Feasibility Rate (\( f_r \)) & \textbf{0.719} & 0.068 & 0.007 \\
Diversity Score (\( D_s \)) & 0.380 & \textbf{0.801} & 0.617 \\
Normalized Temp. (\( \overline{T_{norm}} \)) & \textbf{0.706} & 0.665 & 0.666 \\
Composite Quality Index (CQI) & \textbf{0.1929} & 0.0362 & 0.0029 \\
\hline
\end{tabular}
\end{table}

%%%%%%%%%%%%%%%% end table %%%%%%%%%%%%%%%%%%%

From Table \ref{table_quality_comparison}, it is evident that the Cooling-Guided DDPM outperforms both CTGAN and TabDDPM. The Cooling-Guided DDPM achieves significantly higher feasibility rates, underscoring its robustness in generating feasible cell layouts. Moreover, the average normalized maximum temperature for feasible layouts is also better with the Cooling-Guided DDPM, indicating that it produces more feasible layouts with better cooling efficiency. 

\begin{figure}[t]
\begin{center}
% Adjust the width of the image to suit your needs. For example, \textwidth.
\includegraphics[width = 0.5\textwidth]{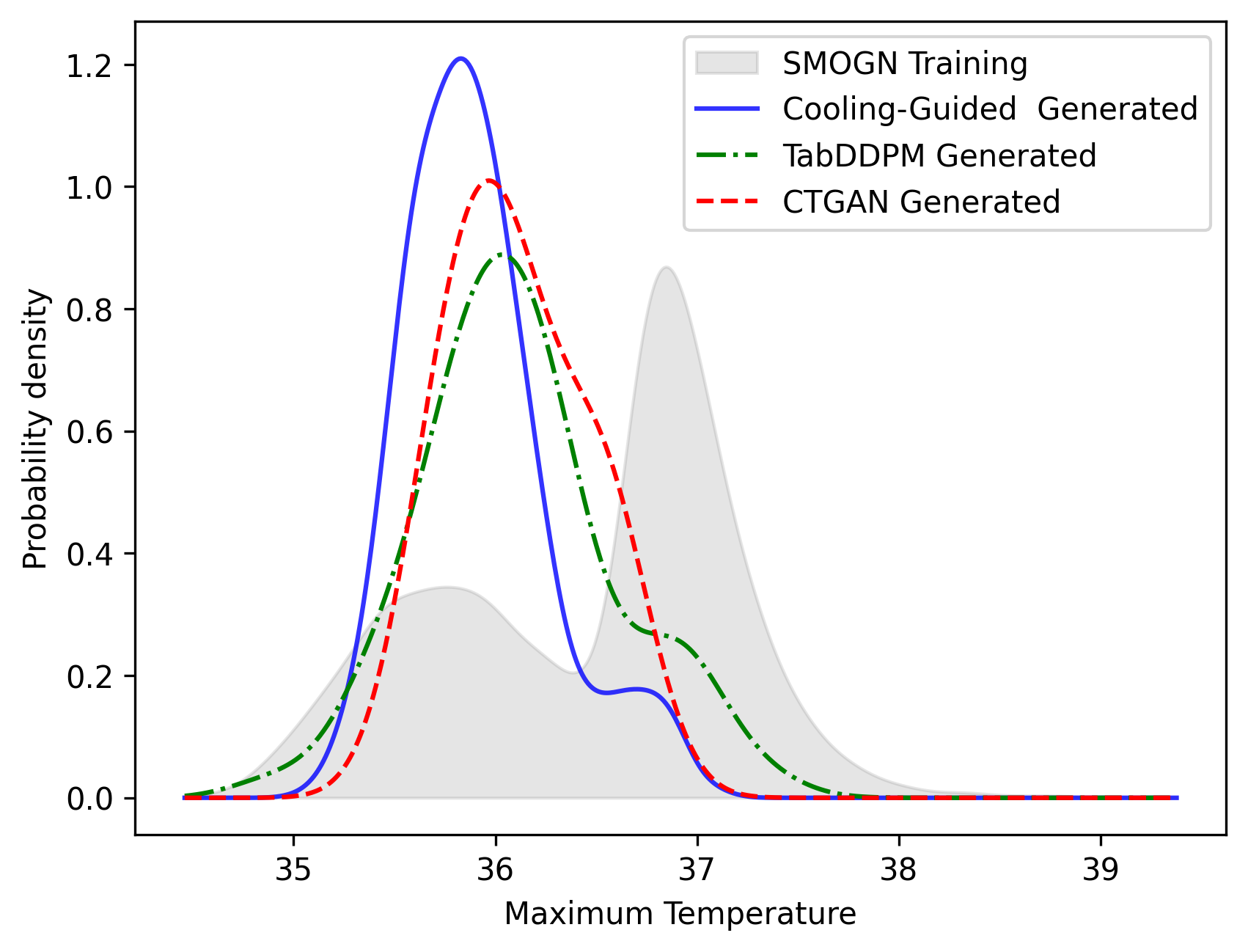}
\end{center}
\caption{COMPARATIVE DISTRIBUTION SHOWING COOLING-GUIDED DDPM GENERATION ACHIEVES LOWEST MAXIMUM TEMPERATURE}
\label{figure_kde} 
\end{figure}

We further explore the distribution of maximum temperatures within the SMOGN-enhanced training data and the layouts generated by the cooling-guided DDPM, TabDDPM, and CTGAN (shown in Figure \ref{figure_kde}). The observed distribution of maximum temperatures in the training data exhibits a bimodal pattern, with significant clusters of samples around 35.8°C and 36.9°C. While all three models successfully approximate the lower mode of this distribution, it is noteworthy that the cooling-guided DDPM demonstrates the best ability to generate samples in the lower temperature spectrum. This indicates a distinct advantage of the cooling-guided DDPM over both the SMOGN-enhanced training data and the other models, showcasing its better ability to produce cell layouts with lower maximum temperatures. 

%%%%%%%%%%%%%%%% begin figure %%%%%%%%%%%%%%%%%%%
%%%%%%%%%%%%%%%% end figure %%%%%%%%%%%%%%%%%%%

While TabDDPM and CTGAN exhibit higher average diversity scores among their limited number of feasible layouts, this metric should be carefully interpreted. The higher diversity score may be because of the limited dataset size, with the scarcity of feasible layouts from TabDDPM and CTGAN potentially inflating the diversity measure rather than indicating a truly comprehensive exploration of the layout space.

To substantiate this interpretation, we conducted a two-dimensional principal component analysis (PCA) on the generated layouts to visualize their spread in comparison to the original training layouts. The PCA, trained on the same training data for all three models, projects the Cooling-Guided DDPM-generated, TabDDPM-generated and CTGAN-generated layouts into a reduced-dimensional space for analysis.

\begin{figure}[t]
\begin{center}
% Adjust the width of the image to suit your needs. For example, \textwidth.
\includegraphics[width = 0.5\textwidth]{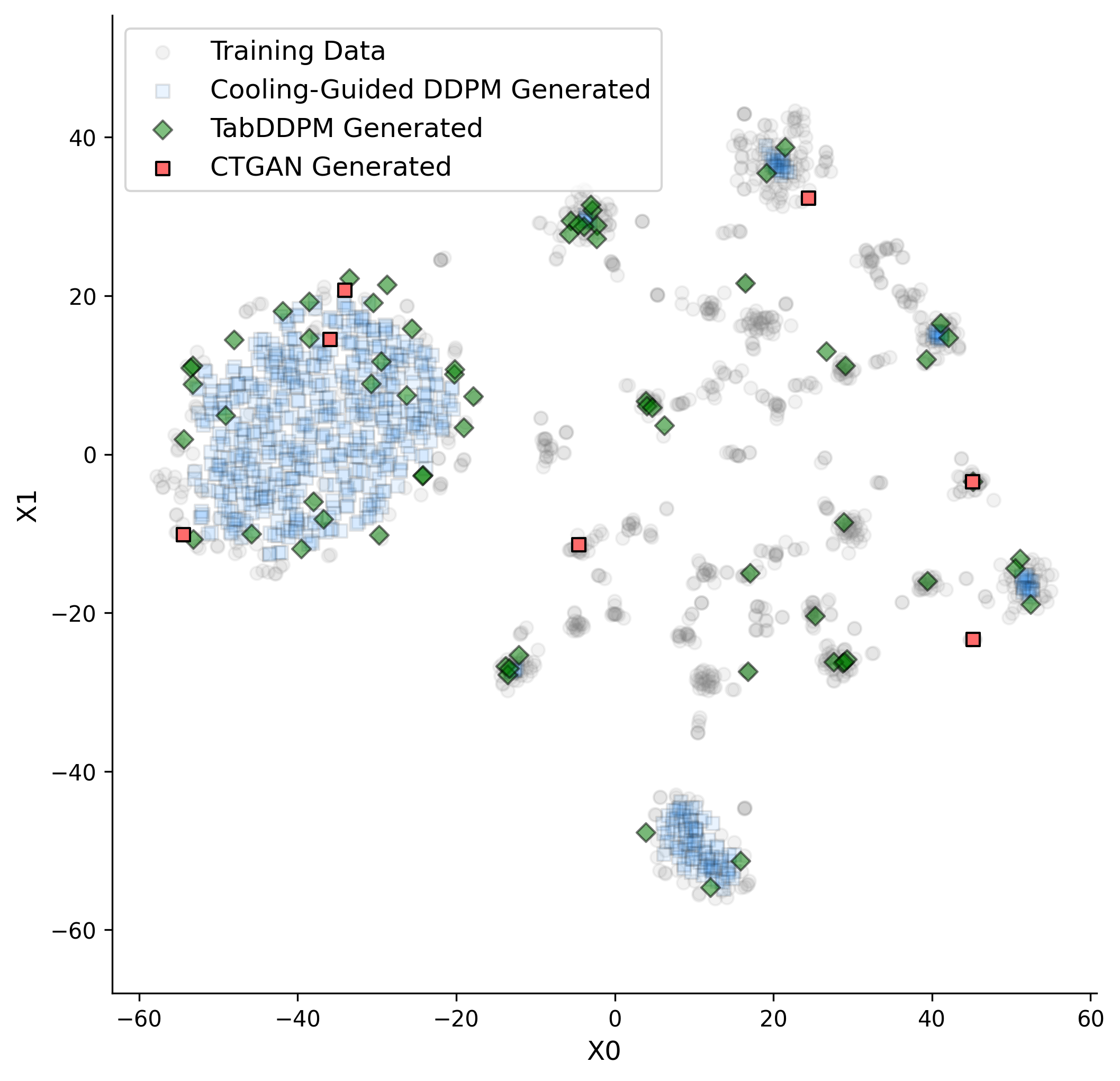}
\end{center}
\caption{2D PCA OF TRAINING AND GENERATED LAYOUTS: COOLING-GUIDED DDPM GENERATION ACHIEVES OPTIMAL LAYOUT DIVERSITY}
\label{figure_tsne} 
\end{figure}

Figure \ref{figure_tsne} demonstrates that the layouts generated by Cooling-Guided DDPM maintain substantial coverage of the original data distribution. More importantly, Cooling-Guided DDPM-generated layouts are not only widespread across the principal component space but also reveal a presence in more clusters than those produced by CTGAN and TabDDPM. This observation substantiates that Cooling-Guided DDPM's lower average diversity score is due to its large number of feasible configurations rather than a lack of variation. This reflects Cooling-Guided DDPM's capability to generate a vast array of diverse, yet feasible layouts which is consistent with the robust and comprehensive coverage of the design space suggested by the training data.

%%%%%%%%%%%%%%%% begin figure %%%%%%%%%%%%%%%%%%%
%%%%%%%%%%%%%%%% end figure %%%%%%%%%%%%%%%%%%%

Overall, the CQI for the Cooling-Guided DDPM is significantly higher than that for TabDDPM and CTGAN, emphasizing the effectiveness of the Cooling-Guided DDPM approach in generating feasible cell layouts with optimal cooling efficiency. Specifically, the CQI for Cooling-Guided DDPM is over 5 times higher than that for TabDDPM and over 66 times higher than that for CTGAN. This result emphasizes the effectiveness of the Cooling-Guided DDPM approach in generating feasible cell layouts with optimal cooling efficiency.

\section*{CONCLUSION}

In this study, we developed a new method for generating battery cell layouts with improved cooling efficiency. Our approach began with the creation of a dataset of 700,000 feasible configurations to train a Denoising Diffusion Probabilistic Model (DDPM). From this dataset, we strategically selected 10,000 configurations for simulation. We found that the distribution was skewed and used the SMOGN library to balance the training data.

This balanced dataset significantly improved the surrogate model's predictions for cell maximum temperatures. Additionally, we generated 700,000 infeasible configurations and combined them with the feasible ones to train a classifier that distinguishes between feasible and infeasible layouts. We then trained a tabular DDPM, conditioned on both the classifier and the surrogate model. This combination of classifier guidance for feasibility and cooling guidance for thermal efficiency optimized the generated layouts.

We established metrics for assessing feasibility, diversity, and cooling efficiency, leading to the development of the Composite Quality Index (CQI) for evaluating the overall quality of generated layouts. After conducting a hyperparameter sweep for \( \lambda \) and \( \gamma \), we determined the optimal settings for achieving the highest CQI. When compared to two state-of-the-art methods, TabDDPM and CTGAN, our approach demonstrated superior performance, with a CQI 5 times greater than that of the TabDDPM and 66 times greater than that of the CTGAN.

This work highlights the effectiveness of integrating generative models with feasibility and thermal performance prediction to generate battery cell layouts optimized for cooling efficiency, offering a significant advancement in battery thermal management system design.

\section*{Acknowledgments}
The authors would like to thank Noah J. Bagazinski for essential base implementations. This work was supported by the Agency for Science, Technology and Research (A*STAR) and the National Science Foundation Engineering Research Center for Power Optimization of Electro-Thermal systems (POETS) with cooperative agreements EEC-1449548.

%Bibliography
\bibliographystyle{unsrt}  
\bibliography{references}

\begin{thebibliography}{10}

\bibitem{jung2015displayed}
Malte~F Jung, David Sirkin, Turgut~M G{\"u}r, and Martin Steinert.
\newblock Displayed uncertainty improves driving experience and behavior: The case of range anxiety in an electric car.
\newblock In {\em Proceedings of the 33rd Annual ACM Conference on Human Factors in Computing Systems}, pages 2201--2210, 2015.

\bibitem{kim2019review}
Jaewan Kim, Jinwoo Oh, and Hoseong Lee.
\newblock Review on battery thermal management system for electric vehicles.
\newblock {\em Applied thermal engineering}, 149:192--212, 2019.

\bibitem{bibin2020review}
Chidambaranathan Bibin, M~Vijayaram, V~Suriya, R~Sai Ganesh, and S~Soundarraj.
\newblock A review on thermal issues in li-ion battery and recent advancements in battery thermal management system.
\newblock {\em Materials Today: Proceedings}, 33:116--128, 2020.

\bibitem{lu2020research}
Mengyao Lu, Xuelai Zhang, Jun Ji, Xiaofeng Xu, and Yongyichuan Zhang.
\newblock Research progress on power battery cooling technology for electric vehicles.
\newblock {\em Journal of Energy Storage}, 27:101155, 2020.

\bibitem{sundin2020thermal}
David~W Sundin and Sebastian Sponholtz.
\newblock Thermal management of li-ion batteries with single-phase liquid immersion cooling.
\newblock {\em IEEE open journal of vehicular technology}, 1:82--92, 2020.

\bibitem{wang2021experimental}
Huaibin Wang, Shuyu Wang, Xuning Feng, Xuan Zhang, Kangwei Dai, Jun Sheng, Zhenyang Zhao, Zhiming Du, Zelin Zhang, Kai Shen, et~al.
\newblock An experimental study on the thermal characteristics of the cell-to-pack system.
\newblock {\em Energy}, 227:120338, 2021.

\bibitem{liu2023generative}
Zheng Liu, Jiaxin Wu, Wuchen Fu, Pouya Kabirazadeh, Sara Kohtz, Nenad Miljkovic, Yumeng Li, and Pingfeng Wang.
\newblock Generative design and optimization of battery packs with active immersion cooling.
\newblock In {\em 2023 IEEE Transportation Electrification Conference \& Expo (ITEC)}, pages 1--5. IEEE, 2023.

\bibitem{li2023machine}
Ao~Li, Jingwen Weng, Anthony Chun~Yin Yuen, Wei Wang, Hengrui Liu, Eric Wai~Ming Lee, Jian Wang, Sanghoon Kook, and Guan~Heng Yeoh.
\newblock Machine learning assisted advanced battery thermal management system: A state-of-the-art review.
\newblock {\em Journal of Energy Storage}, 60:106688, 2023.

\bibitem{kotelnikov2023tabddpm}
A.~Kotelnikov, D.~Baranchuk, I.~Rubachev, and A.~Babenko.
\newblock Tabddpm: Modelling tabular data with diffusion models.
\newblock In {\em International Conference on Machine Learning}, pages 17564--17579, 2023.

\bibitem{ctgan}
Lei Xu, Maria Skoularidou, Alfredo Cuesta-Infante, and Kalyan Veeramachaneni.
\newblock Modeling tabular data using conditional gan.
\newblock In {\em Advances in Neural Information Processing Systems}, 2019.

\bibitem{wang2020cooling}
Haitao Wang, Tao Tao, Jun Xu, Xuesong Mei, Xiaoyan Liu, and Piao Gou.
\newblock Cooling capacity of a novel modular liquid-cooled battery thermal management system for cylindrical lithium ion batteries.
\newblock {\em Applied Thermal Engineering}, 178:115591, 2020.

\bibitem{liu2023control}
Zheng Liu, Jiaxin Wu, Wuchen Fu, Pouya Kabirzadeh, In-Bum Chung, Mohammed~Jubair Dipto, Nenad Miljkovic, Pingfeng Wang, and Yumeng Li.
\newblock Control co-design of battery packs with immersion cooling.
\newblock In {\em ASME International Mechanical Engineering Congress and Exposition}, volume 87592, page V002T02A016. American Society of Mechanical Engineers, 2023.

\bibitem{chen2020immersion}
Pin Chen, Souad Harmand, and Safouene Ouenzerfi.
\newblock Immersion cooling effect of dielectric liquid and self-rewetting fluid on smooth and porous surface.
\newblock {\em Applied Thermal Engineering}, 180:115862, 2020.

\bibitem{yu2022novel}
Xiaoli Yu, Qichao Wu, Rui Huang, and Xiaoping Chen.
\newblock A novel heat generation acquisition method of cylindrical battery based on core and surface temperature measurements.
\newblock {\em Journal of Electrochemical Energy Conversion and Storage}, 19(3):030905, 2022.

\bibitem{ho2020denoising}
Jonathan Ho, Ajay Jain, and Pieter Abbeel.
\newblock Denoising diffusion probabilistic models.
\newblock In {\em Advances in neural information processing systems}, volume~33, pages 6840--6851, 2020.

\bibitem{dhariwal2021diffusion}
P.~Dhariwal and A.~Nichol.
\newblock Diffusion models beat gans on image synthesis.
\newblock In {\em Advances in Neural Information Processing Systems}, volume~34, pages 8780--8794, 2021.

\bibitem{branco2017smogn}
P.~Branco, L.~Torgo, and R.~Ribeiro.
\newblock Smogn: A pre-processing approach for imbalanced regression.
\newblock In {\em Proceedings of Machine Learning Research}, volume~74, pages 36--50, 2017.

\bibitem{bagazinski2023shipgen}
Nicholas~J Bagazinski and Faez Ahmed.
\newblock Shipgen: A diffusion model for parametric ship hull generation with multiple objectives and constraints.
\newblock arXiv preprint arXiv:2306.15166, 2023.

\end{thebibliography}

\end{document}